%% file: root.tex
\documentclass[letterpaper, 10 pt, conference]{ieeeconf}  % Comment this line out if you need a4paper

\IEEEoverridecommandlockouts                              % This command is only needed if
\usepackage[pdftex]{graphicx}
\usepackage{amsmath} % assumes amsmath package installed
\usepackage{amssymb}  % assumes amsmath package installed
\usepackage{subfigure}
\usepackage{multirow}
\usepackage{array,booktabs}
\usepackage{diagbox}
\usepackage{cite}
\usepackage{caption}
\usepackage{amsmath}
\usepackage{hyperref}

\usepackage{irap_SIunits}
\usepackage{irap_acronyms}
\usepackage{irap_math}
\usepackage{irap_misc}

\usepackage{soul,color}

\usepackage{xcolor}
\newcommand{\bl}[1]{{\textcolor{black}{#1}}}

\DeclareMathOperator*{\argmin}{argmin}

\title{\LARGE \bf
Complex Urban LiDAR Data Set
}

\author{Jinyong Jeong${}^{1}$, Younggun Cho${}^{1}$, Young-Sik Shin${}^{1}$, Hyunchul Roh${}^{1}$ and Ayoung Kim${}^{1}$% <-this % stops a space
\thanks{$^{1}$J. Jeong, Y. Cho, Y. Shin, H. Roh and A. Kim are with the Department of Civil and Environmental Engineering, KAIST, Daejeon, S. Korea{ \tt\small [jjy0923, yg.cho, youngsik, rohs\_, ayoungk]@kaist.ac.kr}}%
}

\begin{document}

%\twocolumn[{%
%\renewcommand\twocolumn[1][]{#1}%
%\maketitle
%\begin{center}
%  \centering
%  \includegraphics[width=.98\textwidth]{figs/intro_fig4}
%  \captionof{figure}{
%  This paper provides the complex urban dataset including metropolitan area, apartment building %complex and underground parking lot. \bl{Sample scenes from} the data set can be found in \bl{\href{https://youtu.be/IguZjmLf5V0}{https://youtu.be/IguZjmLf5V0}}.}
%\end{center}%
%}]

\makeatletter
\let\@oldmaketitle\@maketitle% Store \@maketitle
\renewcommand{\@maketitle}{\@oldmaketitle% Update \@maketitle to insert...
\bigskip
\centering
  \includegraphics[width=.98\textwidth]
    {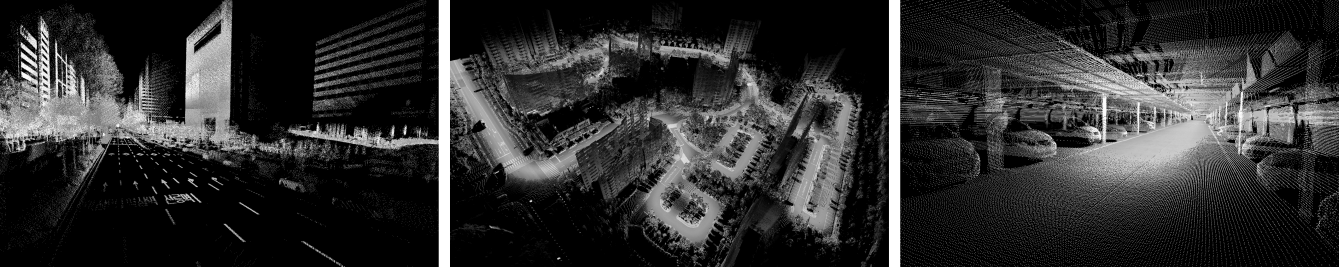}\captionof{figure}{
  This paper provides the complex urban data set including metropolitan area, apartment building complex and underground parking lot. \bl{Sample scenes from} the data set can be found in \bl{\href{https://youtu.be/IguZjmLf5V0}{https://youtu.be/IguZjmLf5V0}}.}  }% ... an image
\makeatother

\maketitle

\renewcommand{\thefigure}{\arabic{figure}}
\setcounter{figure}{1}

\thispagestyle{empty}
\pagestyle{empty}
\input{abstract}

\input{introduction}
\input{related}

\input{system}

\input{data}

\input{conclusion}

\section*{ACKNOWLEDGMENT}

This material is based upon work supported by the \ac{MOTIE}, Korea under
Industrial Technology Innovation Program (No.10051867) and [High-Definition Map Based Precise Vehicle Localization Using Cameras and LIDARs] project funded by Naver Labs Corporation.

\bibliographystyle{IEEEtran}
\bibliography{string-long,references}

\end{document}

%% file: abstract.tex
\begin{abstract}

This \bl{paper} presents a \ac{LiDAR} data set that targets complex urban environments. Urban environments with high-rise buildings and congested traffic pose a significant challenge for many robotics applications. The presented data set is unique in the sense it is able to capture the genuine features of an urban environment (e.g. metropolitan areas, large building complexes and underground parking lots). Data of \ac{2D} and \ac{3D} \ac{LiDAR}, which are typical types of \ac{LiDAR} sensors, are provided in the data set. The two 16-ray \ac{3D} \ac{LiDAR}s are tilted on both sides for maximal coverage. One \ac{2D} \ac{LiDAR} faces backward while the other faces forwards to collect data of roads and buildings, respectively. Raw sensor data from \ac{FOG}, \ac{IMU}, and the \ac{GPS} are presented in a file format for vehicle pose estimation. The pose information of the vehicle estimated at \unit{100}{Hz} is also presented after applying the graph \ac{SLAM} algorithm. For the convenience of development, the file player and data viewer in \ac{ROS} environment were also released via the web page. The full data sets are available at: \bl{\href{http://irap.kaist.ac.kr/dataset}{http://irap.kaist.ac.kr/dataset}}. \bl{In this website, \ac{3D} preview of each data set is provided using WebGL.}

\end{abstract}

%% file: introduction.tex
\section{Introduction} 
\label{sec:intro}

Autonomous vehicles have been studied by many researchers in recent years, and algorithms for autonomous driving have been developed using \bl{diverse} sensors. As it is important that algorithms for autonomous driving use data obtained from the actual environment, many groups have disclosed data sets. Data sets based on camera vision data such as \cite{cordts2016cityscapes}, \cite{brostow2009semantic},
\cite{dollar2009pedestrian} and \cite{geiger2013vision} are used to develop various applications such as visual odometry, semantic segmentation, and vehicle detection. Data sets based on \ac{LiDAR} data such as \cite{geiger2013vision},
\cite{smith2009new}, \cite{blanco2014malaga}, \cite{carlevaris2016university} and
\cite{tong2013canadian} are used in applications such as object detection, \ac{LiDAR} odometry, and 3D mapping. However, most data sets do not focus on highly complex urban environments (significantly wide roads, lots of dynamic objects, \ac{GPS} blackout regions and high-rise buildings) where actual autonomous vehicles operate. 

A complex urban environment such as a downtown area poses a significant challenge for many robotics applications. Validation and implementation in a complex urban environment is not straightforward. Unreliable \ac{GPS}, complex building structure, and limited ground truth are the main challenges for robotics applications in urban environments. In addition, urban environments have high population densities and heavy foot traffic, resulting in many dynamic objects that obstruct robot operations, and cause sudden environmental changes. This paper presents a \ac{LiDAR} sensor data set that specifically targets the urban canyon environment (e.g. metropolitan area and confined building complexes). The data set is not only extensive in terms of time and space, but also includes features of large-scale environments such as skyscrapers and wide roads. The presented data set was collected using two types of \ac{LiDAR}s and various navigation sensors that possess both commercial-level accuracy and high-level accuracy.

\begin{figure}[!b]
  \centering
  \includegraphics[width=0.95\columnwidth] {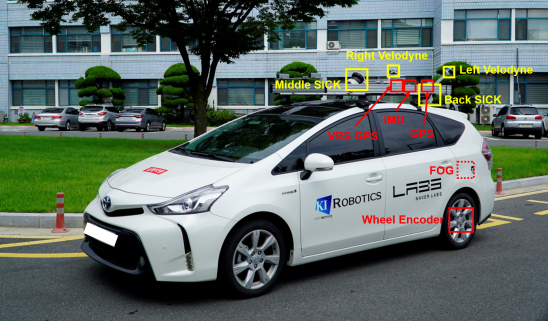}
  \caption{\ac{LiDAR} sensor system for the complex urban data set. The yellow boxes indicate \ac{LiDAR} sensors (2D and 3D \ac{LiDAR} sensors) and the red boxes indicate navigation sensors (\ac{VRS}-\ac{GPS}, \ac{IMU}, \ac{FOG}, and \ac{GPS}).}
  \label{fig:car}
\end{figure}

The structure of the paper is as follows. Section \ref{sec:related} describes the process of surveying existing publicly open data sets and comparing the characteristics. Section \ref{sec:system} provides an overview of the configuration of the sensor system. The details and specificity of the proposed data set are explained in Section \ref{sec:data}. Finally, the conclusion of the study and suggestions for further works are provided in Section \ref{sec:conclusion}.

%% file: related.tex
\section{Related Works} 
\label{sec:related}

%FIGURE
\input{fig_sensor_coord2}

%FIGURE

There are several data sets in the robotics field that offer 3D point cloud data sets of indoor/outdoor environments. The Ford Campus Vision and \ac{LiDAR} Data Set \cite{pandey2011ford} offers 3D scan data of roads and low-rise buildings. The data set was captured in a part of a campus using horizontally scanning 3D LiDAR mounted on the top of a vehicle. The KITTI data set \cite{geiger2013vision} provides \ac{LiDAR} data of \bl{less complex urban areas} and highways, and is the most commonly used data set for various robotic applications including motion estimation, object tracking, and semantic classification. The North Campus Long-Term (NCLT) data set \cite{carlevaris2016university} consists of both 3D and 2D \ac{LiDAR} data collected in the University of Michigan campus. The segway platform explored both indoor and outdoor environments over a period of 15 months to capture long-term data. However, these data sets do not address highly complex urban environments that include various moving objects, high-rise buildings, and unreliable positioning sensor data. 

The Malaga data set \cite{blanco2014malaga} provides 3D point cloud data using two planar 2D \ac{LiDAR} mounted on the side of the vehicle. The sensors were equipped in a push-broom configuration, and 3D point data was acquired as the vehicle moving forward. The Multi-modal Panoramic 3D Outdoor (MPO) data set \cite{jung2016multi} offers two types of 3D outdoor data sets: dense and sparse MPO. This data set mainly focuses on data for semantic place recognition. To obtain dense panoramic point cloud data, the authors utilized a static 3D LiDAR mounted on a moving platform. The Oxford RobotCar (Oxford) Dataset \cite{maddern20171} collected large variations of scene appearance. Similar to the Malaga data set, this data set also used push-broom 2D \ac{LiDAR}s mounted on the front and rear of the vehicle. While the data sets mentioned above attempt to offer various 3D urban information, the data sets are not complex enough to cover the sophisticated environment of complex urban city scenes. 

Compared to these \bl{existing} data sets, the data set presented in this paper possess the following unique characteristics: 

\begin{itemize}
  \item Provides data from diverse environments such as complex metropolitan areas, residential areas and apartment building complexes.

  \item Provides sensor data with two levels of accuracy (economic sensors with consumer-level accuracy and expensive high-accuracy sensors). 

  \item \bl{Provides baseline via \ac{SLAM} algorithm using highly accurate navigational sensors and manual \ac{ICP}.}

  \item Provides development tools for the general robotics community via \ac{ROS}. 

  \item \bl{Provides raw data and \ac{3D} preview using WebGL targeting diverse robot application.}
\end{itemize}

%% file: fig_sensor_coord2.tex
\begin{figure*}[!t]
  \centering
  \begin{minipage}{0.36\textwidth}
  \subfigure[Top view]{%
    \includegraphics[width=0.95\textwidth] {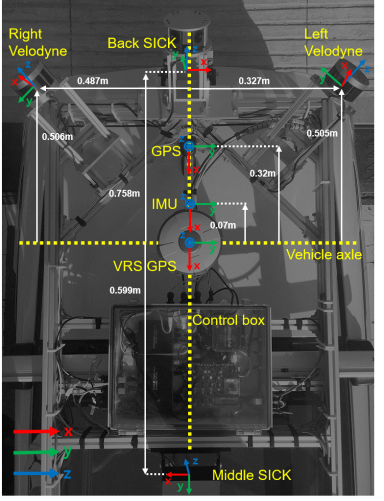}
    \label{fig:sensor_rig_top}
  }
  \end{minipage}
  \begin{minipage}{0.56\textwidth}
  \subfigure[Side view]{%
   	\includegraphics[width=0.95\textwidth] {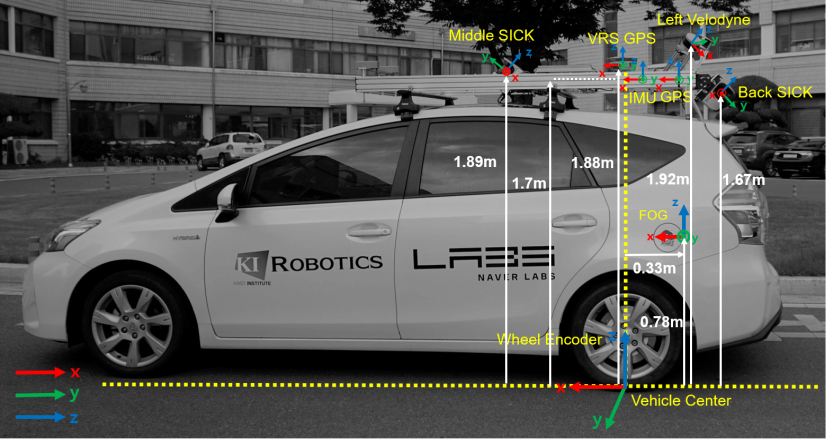}
   	\label{fig:side_view}
  }\\
  \subfigure[Rear view for two 3D LiDARs]{%
   	\includegraphics[width=0.46\textwidth, height=0.2\textwidth] {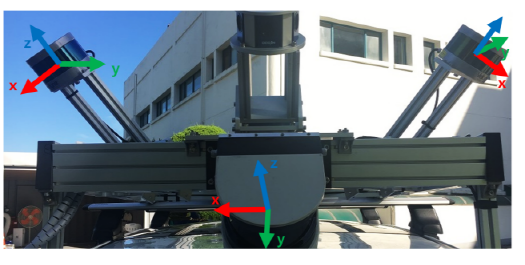}
   	\label{fig:vlp-all-back}
  }
    \subfigure[Side view for two 2D LiDARs]{%
  	\includegraphics[width=0.46\textwidth, height=0.2\textwidth] {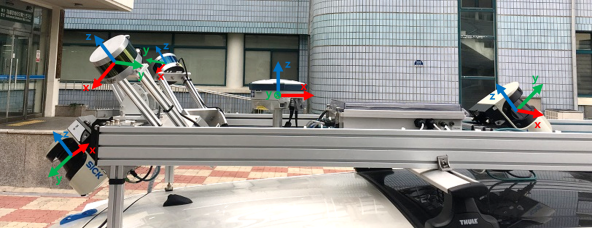}
  	\label{fig:sick_side}
  	}
  \end{minipage}  
  \caption{Hardware sensor configuration. \subref{fig:sensor_rig_top} Top view and \subref{fig:side_view} side view of the entire sensor system with coordinate frame. Each sensor is mounted on the vehicle, and the red, green, and blue arrows indicate the x, y, and z coordinates of the sensors, respectively. \subref{fig:vlp-all-back} Two 3D \ac{LiDAR}s are tilted $45^\circ$ for maximal coverage. \subref{fig:sick_side} The rear 2D \ac{LiDAR} face downwards towards the road and the middle 2D \ac{LiDAR} faces upwards to detect the building structures. Sensor coordinates are displayed on each sensor figure.}
  \label{fig:sensors}
\end{figure*}

\begin{table*}[!t]
  \centering
  \caption{Specifications of sensors used in sensor system (H: Horizontal, V: Vertical)}
  \label{tab:spec}%
  \begin{tabular}{c|ccccccc}%
    Type                   & Manufacturer   & Model         & Description                            		& No.   & Hz    & Accuracy  & Range \\
    \hline
    \textbf{3D LiDAR}      & Velodyne       & VLP-16        & 16 channel 3D LiDAR with 360$^\circ$ FOV         & 2     & \bl{10}    &      & 100 m\\
    \textbf{2D LiDAR}      & SICK           & LMS-511       & 1 channel 2D LiDAR with 190$^\circ$ FOV  		& 2     & 100   &           & 80 m\\
    \textbf{GPS}           & U-Blox         & EVK-7P        & Consumer level GPS                		& 1     & 10    &  2.5 \unit{m}         &\\
    \textbf{VRS GPS}       & SOKKIA         & GRX 2         & VRS-RTK GPS                            		& 1     & 1     &  H: 10 \unit{mm}, V: 15 \unit{mm}   &\\ 
    \textbf{3-axis FOG}    & KVH            & DSP-1760      & Fiber optics gyro (3 axis)             		& 1     & 1000  &  0.05$^\circ$/h  &\\
    \textbf{IMU}           & Xsens          & MTi-300       & Consumer level gyro enhanced AHRS       		& 1     & 100   &  10$^\circ$/h  &\\
    \textbf{Wheel encoder} & RLS            & LM13          & Magnetic rotary encoder                		& 2     & 100   & 4096 (resolution)  &\\
    \textbf{Altimeter}	  & Withrobot      & myPressure    & Altimeter sensor                  		& 1     & 10   & 0.01hPa (resolution)  &\\
    \hline
  \end{tabular}
\end{table*}

%% file: system.tex
\section{System Overview} 
\label{sec:system}

This section describes the sensor configuration of the hardware platform and the sensor calibration method. 

%--------------------------------------------
\subsection{Sensor Configuration}

The main objective of the sensor system in \figref{fig:car} is to provide sensor measurements that possess different sensor accuracy levels. For the attitude and position of the vehicle, data from both relatively low-cost sensors and highly accurate expensive sensors were provided simultaneously. The sensor configuration is summarized in \figref{fig:sensors} and \tabref{tab:spec}. 

The system included both 2D and 3D \ac{LiDAR}s that provide a total of four \ac{LiDAR} sensor measurements. Two 3D \ac{LiDAR}s were installed in parallel facing the rear direction and tilted $45^\circ$ from the longitudinal and lateral planes. The structure of the tilted 3D \ac{LiDAR}s allow for maximal coverage as data on the plane perpendicular to the travel direction of the vehicle can be obtained. Two 2D \ac{LiDAR}s were each installed facing forward and backward, respectively. The rear 2D \ac{LiDAR} faces downwards towards the road, while the frontal \ac{LiDAR} installed in middle portion faces upwards toward the buildings.

For inertial navigational sensors, two types of attitude sensor data, a 3-axis \ac{FOG} and an \ac{IMU}, were provided. The 3-axis \ac{FOG} provides highly accurate attitude measurements that are used to estimate a baseline, while the \ac{IMU} provides general sensor measurements. The system also has two levels of \ac{GPS} sensors, a \ac{VRS} \ac{GPS} and a single \ac{GPS}. The \ac{VRS} \ac{GPS} provides up to cm-level accuracy when a sufficient number of satellites are secured, while the single \ac{GPS} provides conventional-level position measurement. However, note that the availability of \ac{GPS} is limited in urban environments due to the complex environment and the presence of high-rise buildings. 

The hardware configuration for the sensor installation are depicted in \figref{fig:sensors}. \figref{fig:sensor_rig_top} and \figref{fig:side_view} show the top and the side views of the sensor system, respectively. Each sensor possesses its own coordinate system, and the red, green and blue arrows in the figure indicate the x, y and z axes of each coordinate system. The figures also portray the relative coordinate values of each sensor relative to the reference coordinate system of the vehicle. The center of the reference coordinate system is located at the center of the vehicle rear axle with a height of zero. 

Most sensors were mounted externally on the vehicle with the exception of the 3-axis \ac{FOG}, which was installed inside the vehicle as shown. Magnetic rotary encoders were used to gauge wheel rotation, and were installed inside each wheel. The vehicle was equipped with 18-inch tires. All sensor data was logged using a personal computer (PC) with an i7 processor, a 512GB SSD, and 64GB DDR4 memory. The sensor drivers and logger were developed on the Ubuntu OS. Additional details are listed in \tabref{tab:spec}. 

%--------------------------------------------
\subsection{Odometry Calibration}

For accurate odometry measurements, odometry calibration was performed using high-precision sensors: \ac{VRS} \ac{GPS} and \ac{FOG}. The calibration was conducted in a wide and flat open space that guaranteed precision of the reference sensors, \ac{VRS} \ac{GPS} and \ac{FOG}. As two-wheel encoders were mounted on the vehicle, the forward kinematics of the platform can be calculated using three parameters $\mathbf{w} = (d_l, d_r, w_b)$: the left and right wheel diameters, and the wheel base between the two rear wheels. To obtain relative measurements from global motion sensors, a 2D pose graph $\mathbf{x} = (\mathbf{x}_1, \cdots \mathbf{x}_n)$ was constructed whenever accurate \ac{VRS} \ac{GPS} measurements were received. The coordinates of the \ac{VRS} \ac{GPS} and \ac{FOG} are globally synchronized, and a node is added from hard-coupled measurements $\mathbf{x}_i = (x_{i}, y_{i}, \theta_{i})$ \bl{written in vehicle center coordinate}. Least square optimization was used to obtain optimized kinematic parameters $\mathbf{w}^*$ using relative motion from the graph $\mathbf{z}_{i} = \mathbf{x}_{i+1} \ominus
\mathbf{x}_i $ and forward motion from the kinematics $\mathbf{k}_i(\mathbf{w})$. The mathematical expression of the objective function is 

\begin{equation}
  \label{eq:calib_odo}
  \mathbf{w}^* = \argmin_{\mathbf{w}} \sum_i ||\mathbf{z}_i \ominus \mathbf{k}_i(\mathbf{w})||_{\Omega_i}
\end{equation}

where $\ominus$ is the inverse motion operator \cite{smith1990estimating} and $\Omega_i$ represents the measurement uncertainty of \ac{VRS} \ac{GPS} and \ac{FOG}. The calibrated parameters are provided in \texttt{EncoderParameter.txt} file in \texttt{calibration} folder.

%--------------------------------------------
\subsection{LiDAR Extrinsic Calibration}

The purpose of this process is to calculate accurate transformation between the reference vehicle coordinates and the coordinates of each sensor. Three types of extrinsic calibration are required to achieve this purpose. Extrinsic \ac{LiDAR} calibration between the four \ac{LiDAR} sensors was performed via optimization. The \tabref{tab:coord_sub} represents each coordinate frame.

\begin{figure}[!t]
  \centering
  \subfigure[\bl{Front view of \ac{LiDAR} sensor data}]{%
    \includegraphics[width=0.95\columnwidth] {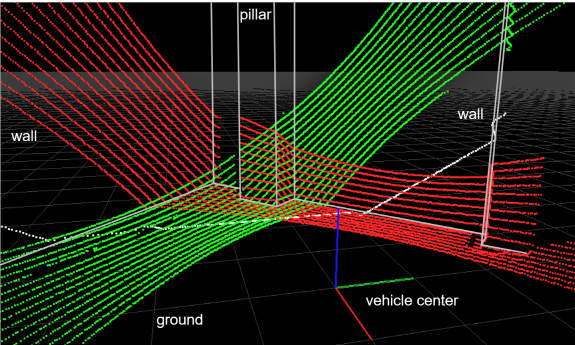}
    \label{fig:calibration1}
  }\\
  \subfigure[\bl{Top view of \ac{LiDAR} sensor data}]{%
    \includegraphics[width=0.95\columnwidth] {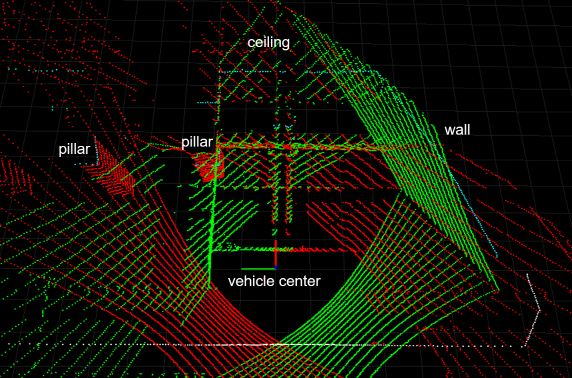}
    \label{fig:calibration2}
  }
  \caption{Point cloud captured during the \ac{LiDAR} calibration. A corner of
  the building was used for the calibration to provide multiple planes
  orthogonal to each other. Red and green point cloud are left and right 3D
  \ac{LiDAR} point cloud respectively. \bl{The white and azure point cloud are rear and middle 2D \ac{LiDAR} point cloud respectively.} The red, green, and blue lines perpendicular to each
  other refer to the reference coordinate system of the vehicle.}
  \label{fig:calibration}
\end{figure}

\begin{table}[!h]
  \centering
  \caption{Coordinate frame subscript}
  \label{tab:coord_sub}%
  \begin{tabular}{c|c}%
    Subscript   & Description\\
    \hline
    $v$ & Vehicle frame\\
    $l$ & Left 3D LiDAR (LiDAR reference frame)\\
    $r$ & Right 3D LiDAR\\
    $m$ & Forward looking 2D LiDAR in the middle\\
    $b$ & backward looking 2D LiDAR in the rear\\
    \hline
  \end{tabular}
\end{table}

\input{tab_data}

%\textbf{\textit{1) 3D LiDAR to 3D LiDAR}} 

\subsubsection{3D LiDAR to 3D LiDAR}

Among the four \ac{LiDAR} sensors installed in the vehicle, the left 3D \ac{LiDAR} sensor was used as a reference frame for calibration. By calculating the relative transformation of other \ac{LiDAR} sensors with respect to the left 3D \ac{LiDAR} sensor, a relative coordinate transform was defined among all the \ac{LiDAR} sensors. The first relative coordinate transform that should be computed is the transform between the left and right 3D \ac{LiDAR} sensors. \ac{GICP}\cite{segal2009generalized} was applied to calculate the required transformation that maps the $i^\text{th}$ right \ac{LiDAR} point cloud data ($\mathbf{p}_{r,i}$) to the corresponding left \ac{LiDAR} point cloud data ($\mathbf{p}_{l,i}$). \figref{fig:calibration} shows the \ac{LiDAR} sensor data \bl{during the calibration process}. As shown in the figure, the relative rotation ($\mathbf{R}_{rl}$) and translation ($\mathbf{t}_{rl}$) of the two 3D \ac{LiDAR} sensors can be calculated using data from the overlap region between the two 3D \ac{LiDAR} data by minimizing the error between the projected points \eqref{eq:calib_3d3d}.

\begin{equation}
  \label{eq:calib_3d3d}
  \mathbf{R}_{rl}^*,\mathbf{t}_{rl}^* = \argmin_{\mathbf{R}_{rl},\mathbf{t}_{rl}} \sum_i \left\{ \mathbf{p}_{l,i} - (\mathbf{R}_{rl} \cdot \mathbf{p}_{r,i} + \mathbf{t}_{rl}) \right\}
\end{equation}

%\textbf{\textit{2) 3D LiDARs to the ground}} 

\subsubsection{3D LiDARs to the Vehicle}

Using the previously computed coordinate transformation, the two 3D \ac{LiDAR} points are aligned to generate merged 3D \ac{LiDAR} points. The next step is to find the transformation to match the ground points in the merged 3D \ac{LiDAR} points to zero. The ground points are first detected using the \ac{RANSAC} algorithm by fitting a plane. The height value of all plane points should be zero. Formulating as above, the least square problem was solved using \ac{SVD}. 

%\begin{equation}
%  \label{eq:calib_ground_transformation}
%   \begin{bmatrix}
%   x_n'\\y_n'\\z_n'\\1
%   \end{bmatrix} = P_{4 \times 4} \begin{bmatrix} 
%   x_n\\y_n\\z_n\\1
%   \end{bmatrix} = \begin{bmatrix}
%   P_1\\P_2\\P_3\\P4
%   \end{bmatrix}_{4 \times 4} \begin{bmatrix} 
%   x_n\\y_n\\z_n\\1
%   \end{bmatrix}\\
%\end{equation}

%\begin{equation}
%  \label{eq:calib_AX0}
%  AX = \begin{bmatrix}
%    x_1 & y_1 & z_1 & 1 \\ x_2 & y_2 & z_2 & 1 \\ \vdots&\vdots&\vdots&\vdots \\ x_n & y_n & z_n & 1
%  \end{bmatrix}
%  P_3^T = 
%  \begin{bmatrix}
%    0\\0\\ \vdots \\ 0
%  \end{bmatrix}
%\end{equation}

%In \eqref{eq:calib_ground_transformation}, $P$ is a $4\times4$ transformation
%matrix for matching the height of ground points extracted from the merged 3D
%points to zero. Since the height ($z$) of points after transformation is
%affected only by the third row ($P_3$) of the transformation matrix ($P$), it
%becomes a problem to calculate $P_3$. If an $A$ matrix is created using ground
%points as in \eqref{eq:calib_AX0}, it is a problem to calculate X ($P_3$)
%satisfying AX = 0. Therefore, in the \ac{SVD} result of $A$ matrix, the last
%column of $V$ becomes the solution of $P_3$, and the rotation matrix and $z$
%translation value can be calculated from $P_3$.

%\textbf{\textit{3) 3D LiDAR to 2D LiDAR}} 

\subsubsection{3D LiDAR to 2D LiDAR}

Completing the previous two steps calculates the transformation between the vehicle and the two 3D \ac{LiDAR} coordinates and the resulting point cloud is properly grounded. In the following step, 3D \ac{LiDAR} data that overlap with 2D LiDAR data are used to estimate the transformation between the 2D \ac{LiDAR} sensor and the vehicle. Structural information was used to consider plane-to-point alignment. Planes are extracted from 3D \ac{LiDAR} data and points from the 2D scan lines are examined in this optimization process. Through this process, it is possible to calculate the transformation from the vehicle to each 2D \ac{LiDAR} sensor ($\mathbf{P}_{vb}, \mathbf{P}_{vm}$). 

\begin{eqnarray}
  \mathbf{P}_{vb} &=& \mathbf{P}_{vl} \times \mathbf{P}_{bl}^{-1}\nonumber\\
  \mathbf{P}_{vm} &=& \mathbf{P}_{vl} \times \mathbf{P}_{ml}^{-1}
\end{eqnarray}

For accurate calibration values, the transformation was provided in both $SE(3)$ and Euler formats with the data set. \tabref{tab:calib_param} shows calibrated sample coordinate transforms.

\begin{table}[!h]
  \centering
  \caption{Summary of \ac{LiDAR} sensor transformation. Positional data is in meter and rotational data is in degree.}
  \label{tab:calib_param}%
  \begin{tabular}{c|l}%
    Type        & Description [x, y, z, roll, pitch, yaw]\\
    \hline\hline
    $\mathbf{P}_{vl}$    & Vehicle \bl{w.r.t} left 3D LiDAR\\
                & $[-0.505, 0.327, 1.926, 1.618^\circ, 44.84^\circ, 137.0^\circ]$\\
    \hline
    $\mathbf{P}_{vr}$    & Vehicle \bl{w.r.t} \bl{right} 3D LiDAR\\
                & $[-0.506, -0.488, 1.939, -179.5^\circ, 135.7^\circ, 46.02^\circ]$\\
    \hline
    $\mathbf{P}_{vb}$ 	& Vehicle \bl{w.r.t} \bl{rear} 2D LiDAR\\
                & $[-0.758, -0.107, 1.673, 46.08^\circ, -179.8^\circ, 90.30^\circ]$\\
    \hline
    $\mathbf{P}_{vm}$  	& Vehicle \bl{w.r.t} middle 2D LiDAR\\
                & $[0.600, -0.111, 1.890, 145.5^\circ, 1.371^\circ, 90.11^\circ]$\\
    \hline
  \end{tabular}
\end{table}

%% file: tab_data.tex
\begin{table*}[!t]
  \centering
  \caption{Dataset lists}
  \label{tab:datalist}%
  \begin{tabular}{c|ccccccc}%
    Data number    & No. Subset        & Location      &Description   & GPS reception rate    & Complexity  & Wide road rate & Path length \\
    \hline\hline
    Urban00        &      2  			& Gangnam, Seoul & Metropolitan area   & 7.49 & $\bigstar\bigstar\bigstar$ & $\bigstar\bigstar\bigstar\bigstar\bigstar$   & 12.02 km\\
    Urban01        &      2 			& Gangnam, Seoul & Metropolitan area  & 5.3 & $\bigstar\bigstar\bigstar$ & $\bigstar\bigstar\bigstar\bigstar$  & 11.83 km\\
    Urban02        &      2 			& Gangnam, Seoul & Residential area  & 4.58 & $\bigstar\bigstar\bigstar\bigstar\bigstar$ & $\bigstar$   & 3.02 km\\
    Urban03        &      1 			& Gangnam, Seoul & Residential area & 4.57 & $\bigstar\bigstar\bigstar\bigstar\bigstar$ & $\bigstar$   & 2.08 km\\
    \hline
    Urban04        &      3 			& Pangyo & Metropolitan area & 7.31 & $\bigstar\bigstar\bigstar$ & $\bigstar\bigstar\bigstar\bigstar$  & 13.86 km\\
    Urban05        &      1 			& Daejeon          & Apartment complex & 7.56 & $\bigstar\bigstar$ & $\bigstar\bigstar$   & 2.00 km\\
    \hline
  \end{tabular}
\end{table*}

%% file: data.tex
\section{Complex Urban Data Set} 
\label{sec:data}

This section describes the urban \ac{LiDAR} data set regarding formats, sensor types and development tools. The data provides a diverse level of complexity captured in a real urban environment.

%--------------------------------------------
\subsection{Data Description}

The data set of this paper covers various features in large urban areas from wide roads with ten lanes or greater, to substantially narrow roads with high-rise buildings. \tabref{tab:datalist} describes the overview of the data set. As the data set covers highly complex urban environments where \ac{GPS} is sporadic, the depicted \ac{GPS} availability map was overlaid on the mapping route as in \figref{fig:gps_fig}. \tabref{tab:datalist} shows the \ac{GPS} reception rate, which represents the average number of satellites of \ac{VRS} \ac{GPS} data, for each data set. Ten satellites are required to calculate the accurate location regularly. The complexity and wide road rate were evaluated for each data set and are shown in \tabref{tab:datalist}.

\input{fig_route}

%--------------------------------------------
\subsection{Data Format}

\begin{figure}[!t]
  \centering
  \includegraphics[width=0.9\columnwidth] {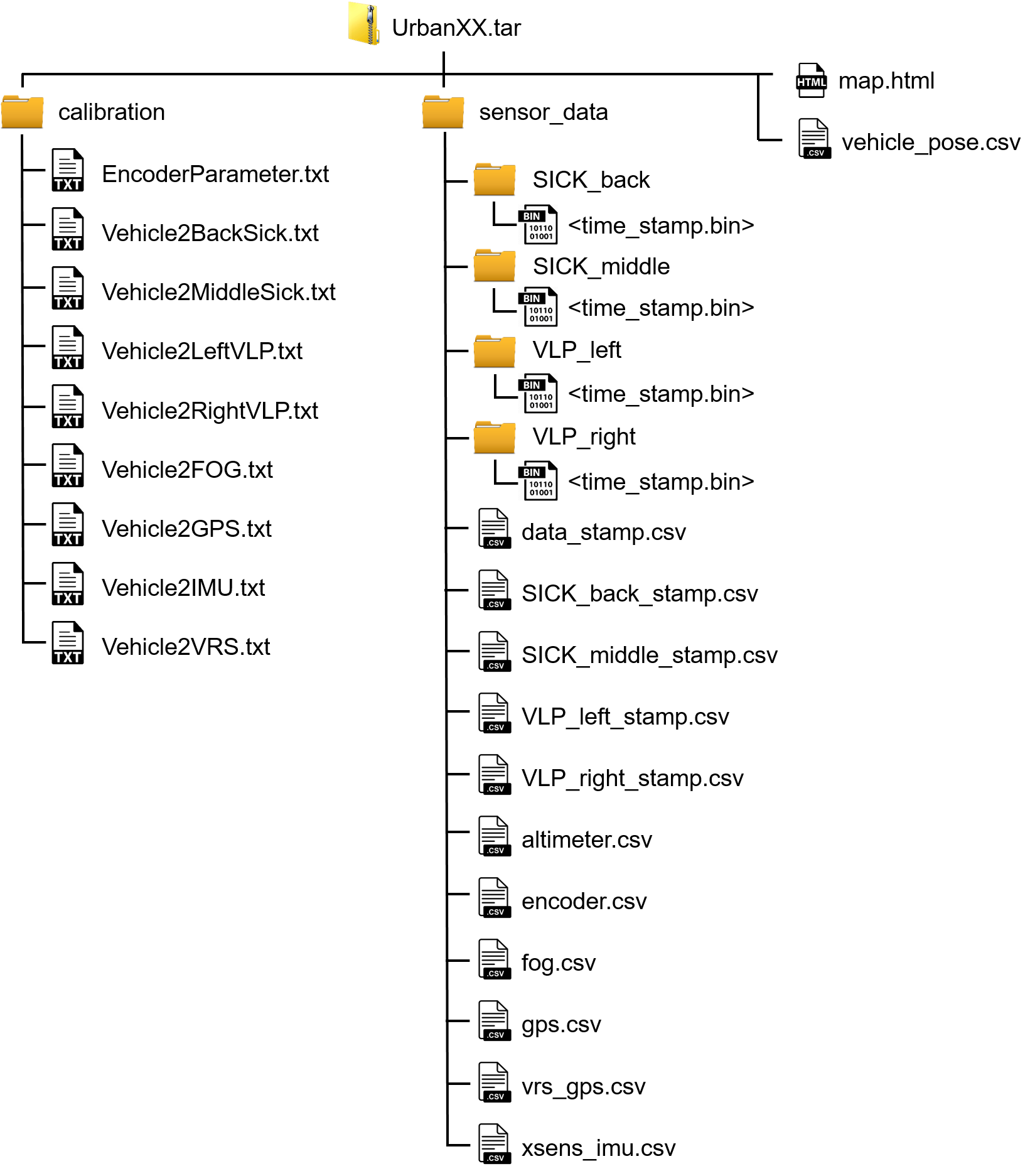}   
  \caption{File directory layout for a single data set.}
  \label{fig:file_directory}
\end{figure}

For convenience in downloading the data, the entire data was split into subsets of approximately \unit{6}{GB} in size. Both the whole data set and the subsets are provided. The path of each data set can be checked through the \texttt{map.html} file in each folder. The file structure of each data set is depicted in \figref{fig:file_directory}. All data was logged using \ac{ROS} timestamps. The data set is in a compressed \texttt{tar} format. For accurate sensor transformation values, calibration was performed prior to each data acquisition. The corresponding calibration data can be found in the \texttt{calibration} folder along with the data. All sensor data is stored in the \texttt{sensor$\_$data} folder.

\begin{enumerate}

%------------------------------%
\item 3D \ac{LiDAR} data

The 3D \ac{LiDAR} sensor, Velodyne (VLP-16), provides data on a per-packet basis. Velodyne’s rotation rate is \unit{10}{Hz}, and the timestamp of the last packet is used as the timestamp of the data at the end of one rotation. 3D \ac{LiDAR} data is stored in the \texttt{VLP$\_$left} and \texttt{VLP$\_$right} folders in the \texttt{sensor$\_$data} folder in a floating-point binary format, and the timestamp of each rotation data is used as the name of the file (\texttt{<time$\_$stamp.bin>}). Each point consists of four items ($x$, $y$,
$z$, $R$). $x$, $y$, and $z$ denote the local 3D Cartesian coordinate values of each \ac{LiDAR} sensor, and $R$ is the reflectance value. The timestamps of all 3D \ac{LiDAR} data are stored sequentially in \texttt{VLP$\_$left$\_$stamp.csv} and \texttt{VLP$\_$right$\_$stamp.csv}. 

%------------------------------%
\item 2D \ac{LiDAR} data

In the system, the 2D \ac{LiDAR} sensors were operated at \unit{100}{Hz}. The 2D \ac{LiDAR} data is stored in the \texttt{SICK$\_$back} and the \texttt{SICK$\_$middle} folder in the \texttt{sensor$\_$data} folder in a floating-point binary format. Similar to 3D \ac{LiDAR} data, the timestamp of each scan data is used as the name of the file. To reduce the file size, the data of 2D \ac{LiDAR} consists of two items ($r$, $R$). $r$ is the range value of each point, and $R$ is the reflectance value. The sensor’s \ac{FOV} is $190^\circ$, where the start angle of the first data is $-5^\circ$, and the end angle is $185^\circ$. The angle difference between each sequential data is $0.666^\circ$. \bl{Each point can be converted from $i^\text{th}$ range measurement to a Cartesian coordinate using this information \eqref{eq:2D_lidar}}.
The timestamps of all 2D \ac{LiDAR} data are stored sequentially in \texttt{SICK$\_$back$\_$stamp.csv} and \texttt{SICK$\_$middle$\_$stamp.csv}. 

\begin{eqnarray}
  \label{eq:2D_lidar}
  x_i &= r \cos(-5 + i \times 0.666)\nonumber\\ 
  y_i &= r \sin(-5 + i \times 0.666)
\end{eqnarray}

%------------------------------%
\item Data sequence

The \texttt{sensor$\_$data/data$\_$stamp.csv} file stores the names and
timestamps of all sensor data in order in the form of (timestamp, sensor name).

%------------------------------%
\item Altimeter data

The \texttt{sensor$\_$data/altitude.csv} file stores the altitude values
measured by the altimeter sensor in the form of (timestamp, altitude).

%------------------------------%
\item Encoder data

The \texttt{sensor$\_$data/encoder.csv} file stores the incremental pulse count
values of the wheel encoder in the form of (timestamp, left count, right count).

%------------------------------%
\item \ac{FOG} data

The \texttt{sensor$\_$data/fog.csv} file stores the relative rotational motion
between consecutive sensor data in the form of (timestamp, delta roll, delta
pitch, delta yaw).

%------------------------------%
\item \ac{GPS} data

The \texttt{sensor$\_$data/gps.csv} file stores the global position measured by
commercial level \ac{GPS} sensor. The data format is (timestamp, latitude,
longitude, altitude, 9-tuple vector (position covariance)).

%------------------------------%
\item \ac{VRS} \ac{GPS} data

The \texttt{sensor$\_$data/vrs$\_$gps.csv} file stores the accurate global
position measured by \ac{VRS} \ac{GPS} sensor. The data format is (timestamp,
latitude, longitude, x coordinate, y coordinate, altitude, fix state, number of
satellite, horizontal precision, latitude std, longitude std, altitude std,
heading validate flag, magnetic global heading, speed in knot, speed in km,
GNVTG mode). \bl{The x and y coordinates use the UTM coordinate system in the meter
unit.} The fix state is a number
indicating the state of the \ac{VRS} \ac{GPS}. For example, 4, 5, and 1
indicates the fix, float, and normal states, respectively. The accuracy of the
\ac{VRS} \ac{GPS} in the sensor specification list (\tabref{tab:spec}) is the
value at the fix state.

%------------------------------%
\item \ac{IMU} data

The \texttt{sensor$\_$data/imu.csv} file stores the incremental rotational pose
data measured by AHRS \ac{IMU} sensor. The data format is (timestamp, quaternion
x, quaternion y, quaternion z, quaternion w, Euler x, Euler y, Euler z).

\end{enumerate}

\begin{figure}[!b]
  \centering
  \includegraphics[width=0.95\columnwidth] {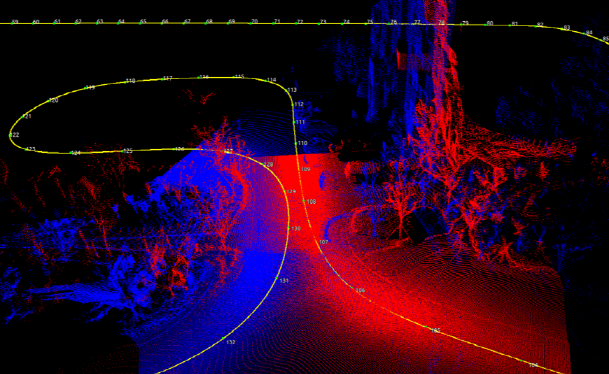}
  \caption{\bl{Baseline generation process using ICP. The yellow line is the path of the vehicle and the number with the green point is the number of the graph node. The red and blue pointcloud are local sub-map of two nodes. The relative poses of the nodes are computed by ICP.}}
  \label{fig:baseline}
\end{figure}

\input{fig_sample2}

%--------------------------------------------
\subsection{Baseline Trajectory using SLAM}

The most challenging issue regarding the validity of data sets is to obtain a reliable baseline trajectory under highly sporadic \ac{GPS} measurements. Both consumer-level \ac{GPS} and \ac{VRS} \ac{GPS} suffer from \ac{GPS} blackouts due to building complexes. 

\bl{In this study, baselines were generated via pose-graph \ac{SLAM}. Our
strategy is to incorporate highly accurate sensors (VRS GPS, FOG and wheel
encoder) in the initial baseline generation. Further refinement over this
initial trajectory is performed using semi-automatic \ac{ICP} for the revisited
places (\figref{fig:baseline}). Manual selection of the loop-closure proposal is
piped into the \ac{ICP} as the initial guess, and the baseline trajectory is refined
using ICP results as the additional loop-closure constraint.}

%\bl{\ac{GPS} reception within our dataset is highly sporadic, and the status
%\ac{GPS} availability in a node changes from fixed state to floating. As an
%evaluation, we examined the subset of the baseline nodes only when the received
%GPS is at the fixed state (2 \unit{cm} precision). For these nodes, we computed
%the root mean squared position error of the estimated baseline graph node
%against \ac{VRS} \ac{GPS} measurement. The average residual over this subset is
%very subtle ($5.5 \cross 10^{-15}$).}

The generated baseline trajectory is stored in \texttt{vehicle$\_$pose.csv} at
the rate of \unit{100}{Hz}. However, it is not desirable to use baseline
trajectory as the ground truth for mapping or localization benchmarking as the
SLAM results depend on the complexity of the urban environment.

%--------------------------------------------
\subsection{Development Tool}

The following tools were provided for the robotics community along with the data set.

\begin{enumerate}

\item File player

To support the \ac{ROS} community, a File player that publishes sensor data as \ac{ROS} messages is provided. New message types were redefined to convey more information, and were released via the GitHub webpage. In urban environments, there are many stop periods during data logging. As most of the algorithm does not require data in stop periods, the player can skip the stop period for convenience, and control data publishing speeds.

\item Data viewer

A data viewer is provided to check the data transmitted through the file player. The data viewer allows users to monitor the data that the player publishes in a visual manner. \bl{The data viewer shows all sensor data and the 2D and 3D \ac{LiDAR} data converted to the vehicle coordinate system.} The Provided player and viewer were built with libraries provided by \ac{ROS} without additional dependencies.

\end{enumerate}

%% file: fig_route.tex
\begin{figure*}[!t]
  \centering
  \subfigure[Urban00 (Gangnam, Seoul, Metropolitan area)]{%
    \includegraphics[clip, trim = 0cm 3cm 0cm 7cm, width=0.32\textwidth ,height=0.20\textwidth] {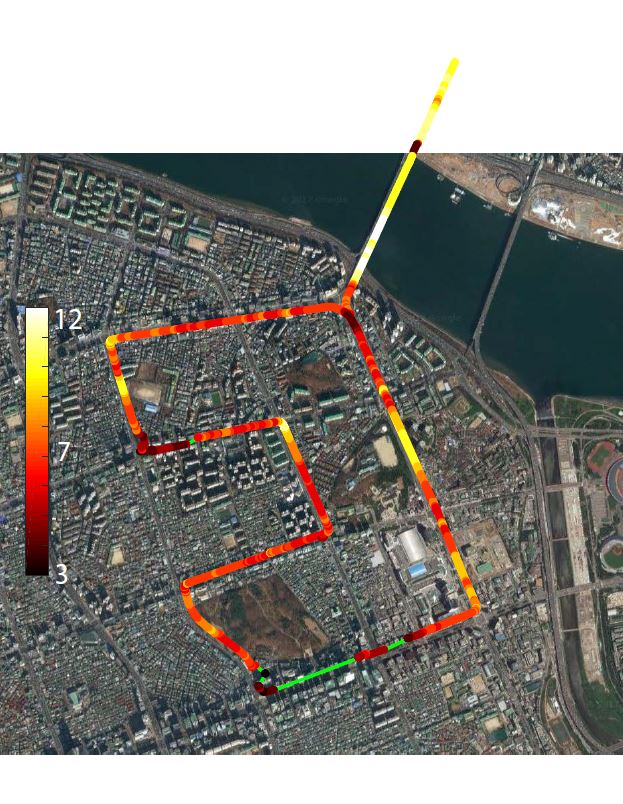}
    \label{fig:gps_seoul1}
  }
  \subfigure[Urban01 (Gangnam, Seoul, Metropolitan area)]{%
    \includegraphics[clip, trim = 0cm 8.5cm 0cm 9cm, width=0.32\textwidth ,height=0.20\textwidth] {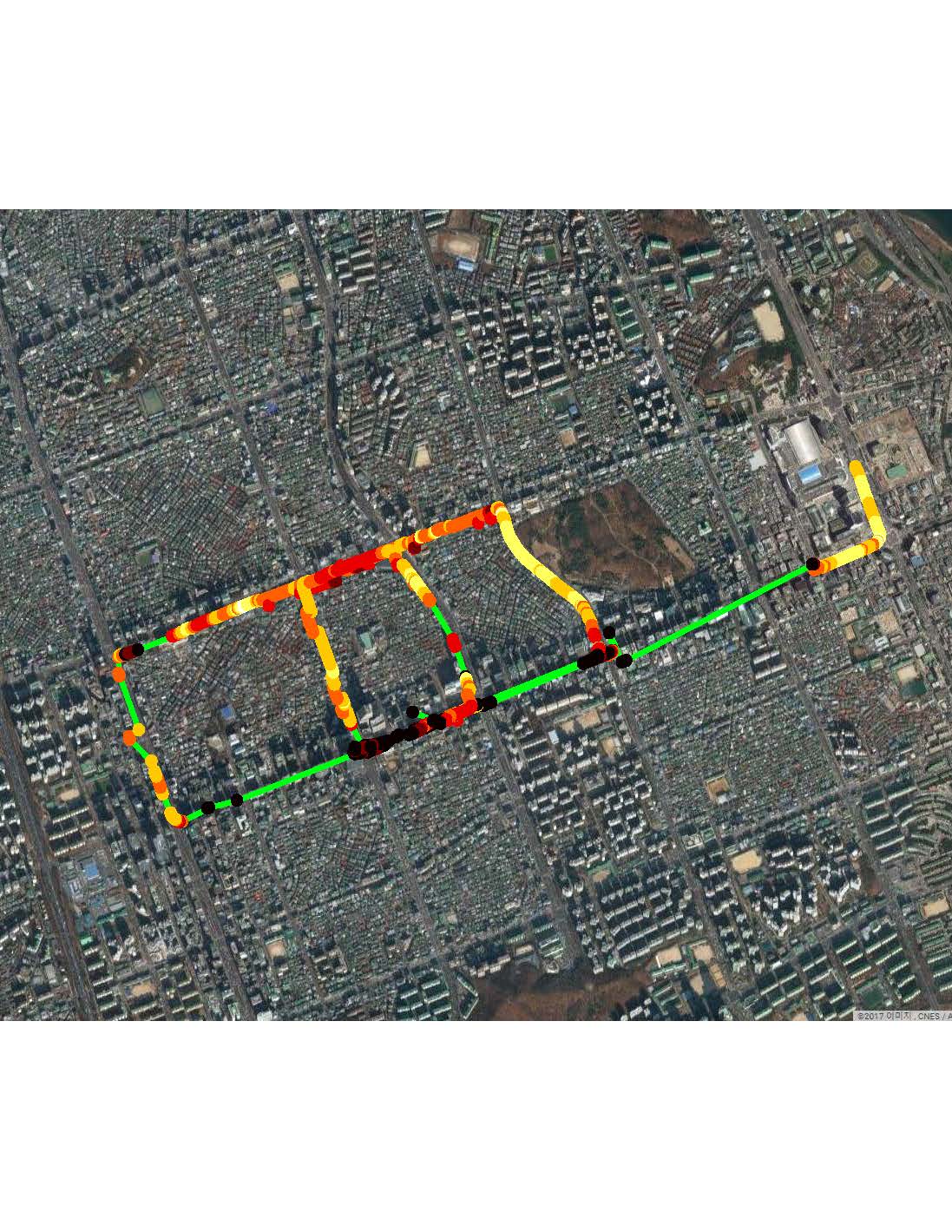}
    \label{fig:gps_seoul2}
  } 
  \subfigure[Urban02 (Gangnam, Seoul, Residential area)]{%
    \includegraphics[clip, trim = 0cm 2cm 0cm 2cm, width=0.32\textwidth ,height=0.20\textwidth] {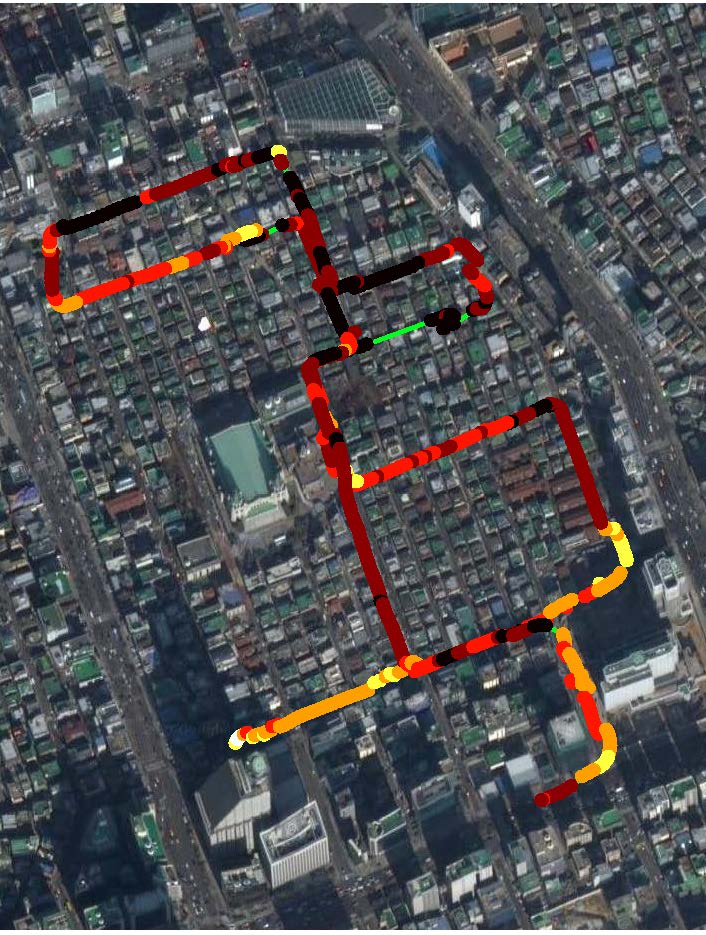}
    \label{fig:gps_seoul3}
  }\\  
  \subfigure[Urban03 (Gangnam, Seoul, Residential area)]{%
    \includegraphics[clip, trim = 0cm 7cm 0cm 6cm, width=0.32\textwidth ,height=0.20\textwidth] {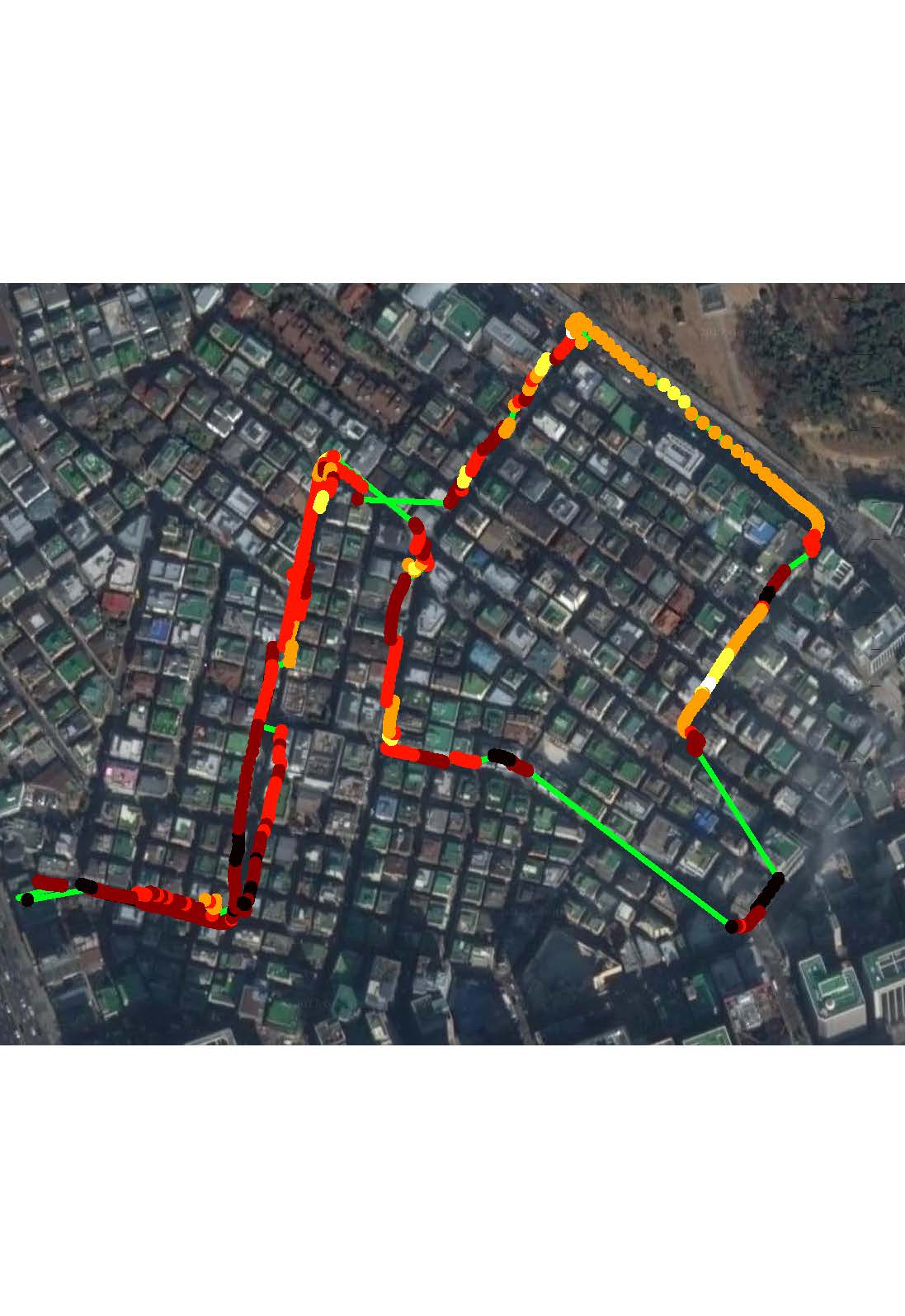}
    \label{fig:gps_seoul4}
  } 
  \subfigure[Urban04 (Pangyo, Metropolitan area)]{%
    \includegraphics[clip, trim = 0cm 5cm 0cm 5cm, width=0.32\textwidth ,height=0.20\textwidth] {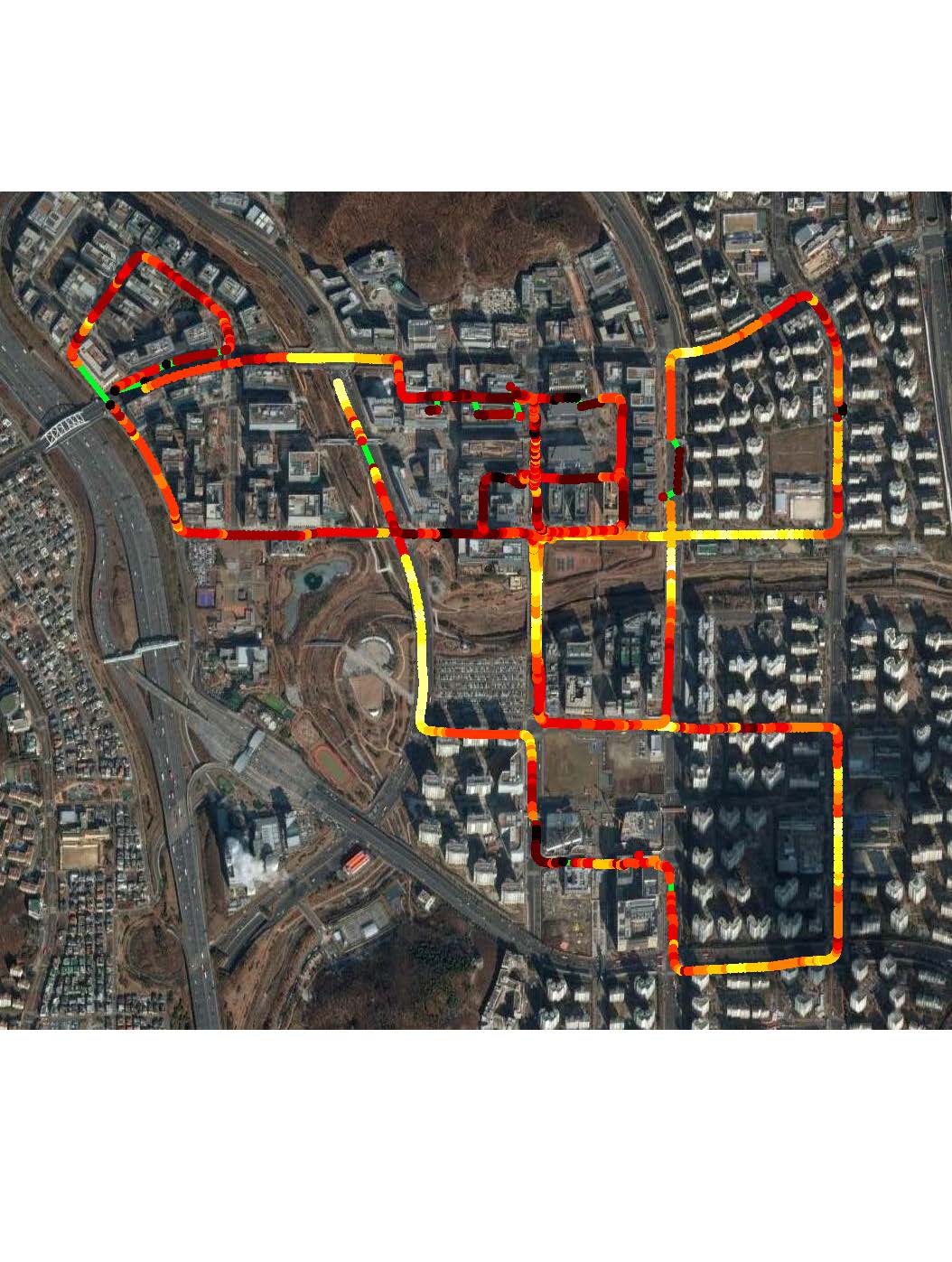}
    \label{fig:gps_pankyo}
  }
  \subfigure[Urban05 (Daejeon, Apartment complex)]{%
    \includegraphics[clip, trim = 0cm 5cm 0cm 5cm, width=0.32\textwidth ,height=0.20\textwidth] {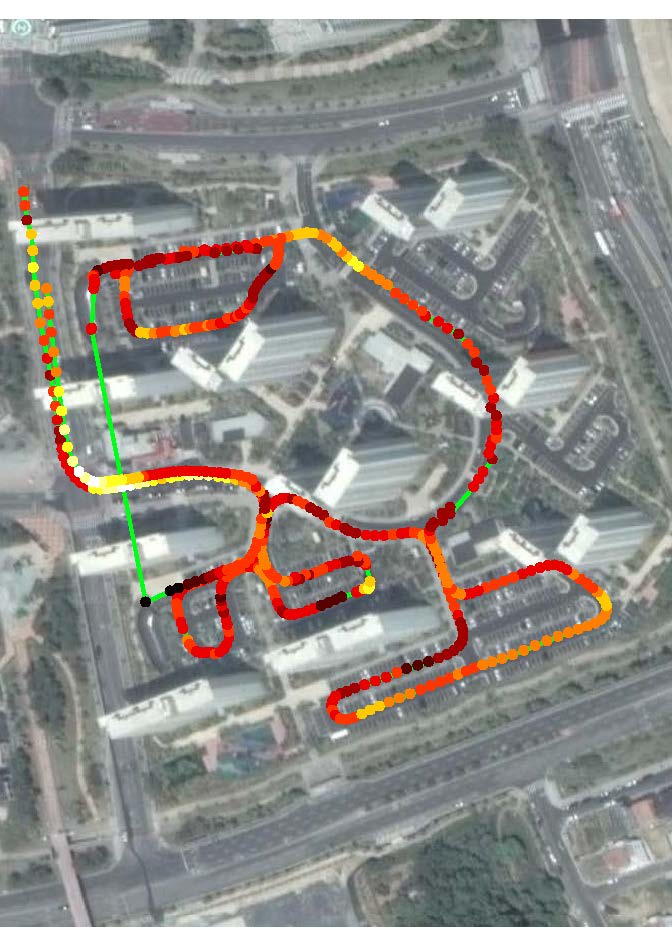}
    \label{fig:gps_doan}
  }\\
  \caption{Data collection route illustrating \ac{VRS} \ac{GPS} data. The green line represents the \ac{VRS} \ac{GPS} based vehicle path. The color of the circles drawn in the route represents the number of satellites used in the \ac{GPS} calculation result; a brighter circle indicates more satellites were used. As complexity increases, fewer satellites are seen. The sections without circles are the areas where no satellites are seen, and no solution of position is available.}
  \label{fig:gps_fig}
\end{figure*}

%% file: fig_sample2.tex
\begin{figure*}[!t]
  \centering
  \subfigure[Wide road (3D \ac{LiDAR})]{%
    \includegraphics[width=0.46\textwidth ,height=0.26\textwidth] {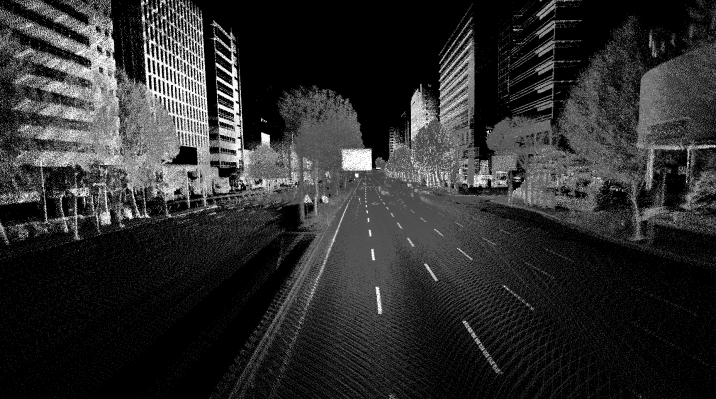}
    \label{fig:data_sample1}
  }
  \subfigure[Wide road (2D \ac{LiDAR})]{%
    \includegraphics[width=0.46\textwidth ,height=0.26\textwidth] {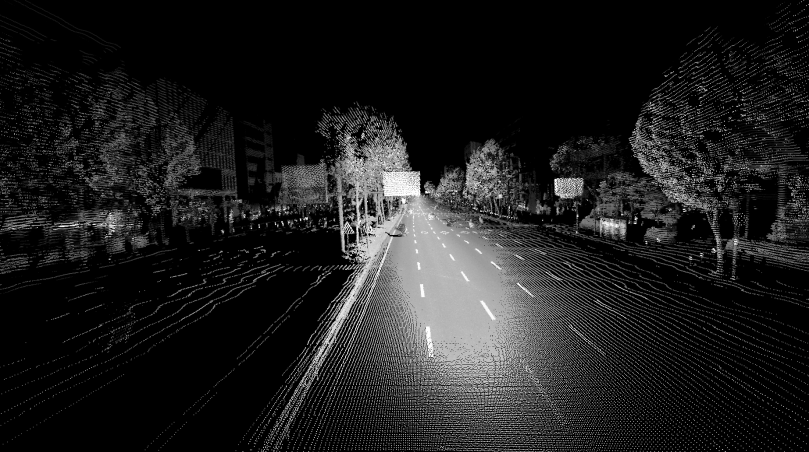}
    \label{fig:data_sample1}
  }\\  
  \subfigure[Complex building enterance (3D \ac{LiDAR})]{%
    \includegraphics[width=0.46\textwidth ,height=0.26\textwidth] {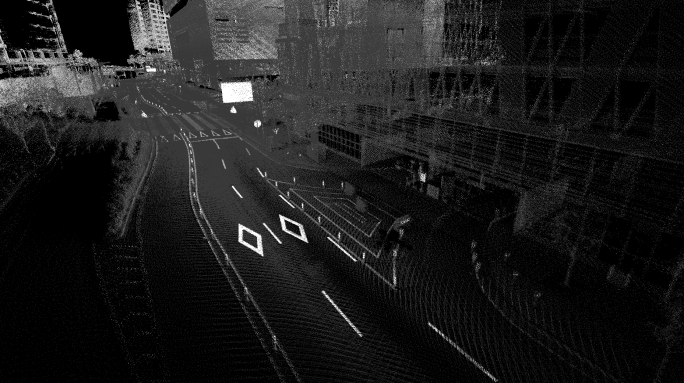}
    \label{fig:data_sample1}
  }
  \subfigure[Complex building enterance (2D \ac{LiDAR})]{%
    \includegraphics[width=0.46\textwidth ,height=0.26\textwidth] {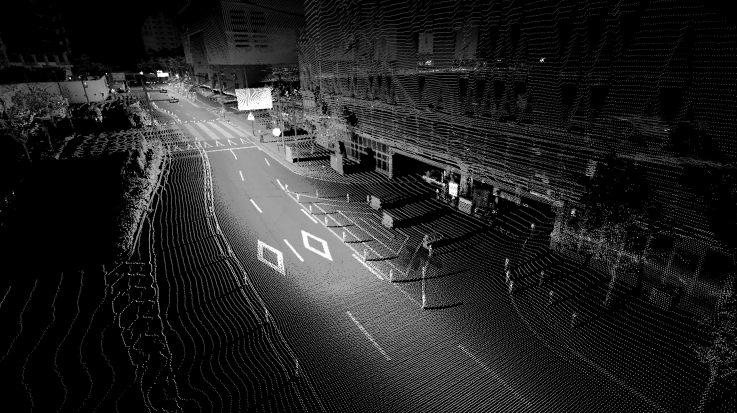}
    \label{fig:data_sample1}
  } \\
  \subfigure[Road markings (3D \ac{LiDAR})]{%
    \includegraphics[width=0.46\textwidth ,height=0.26\textwidth] {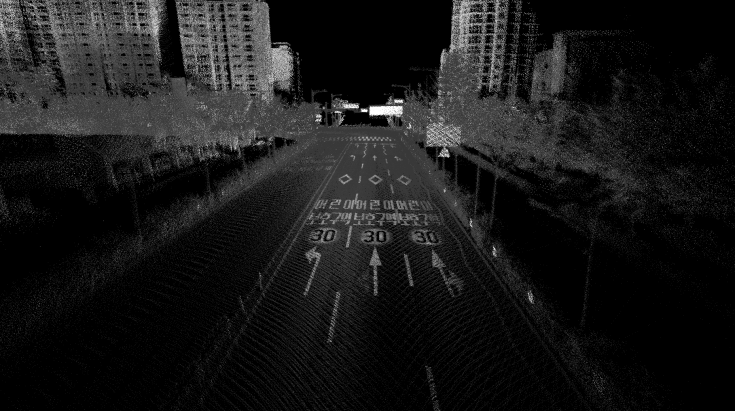}
    \label{fig:data_sample1}
  }
  \subfigure[Road markings (2D \ac{LiDAR})]{%
    \includegraphics[width=0.46\textwidth ,height=0.26\textwidth] {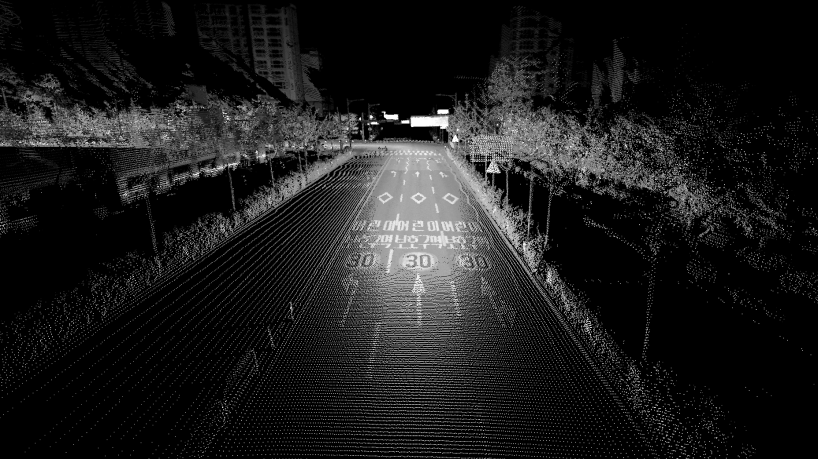}
    \label{fig:data_sample1}
  }\\
  \subfigure[High-rise buildings in complex urban environment (3D \ac{LiDAR})]{%
    \includegraphics[width=0.46\textwidth ,height=0.26\textwidth] {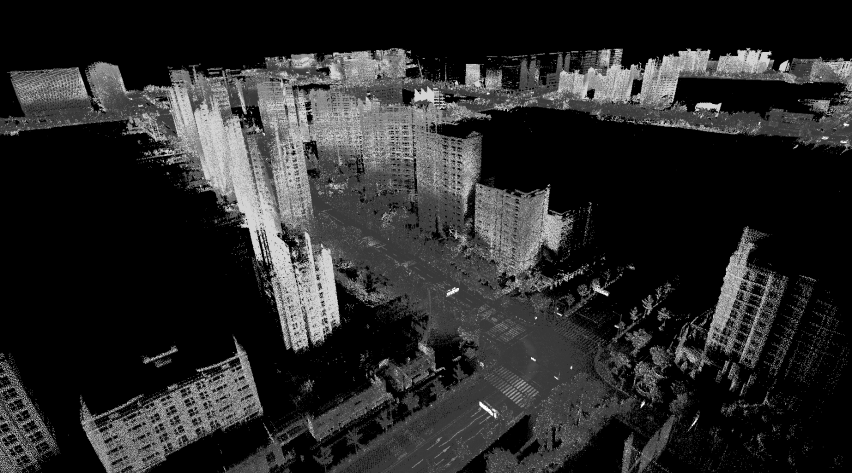}
    \label{fig:data_sample1}
  }
  \subfigure[High-rise buildings in complex urban environment (2D \ac{LiDAR})]{%
    \includegraphics[width=0.46\textwidth ,height=0.26\textwidth] {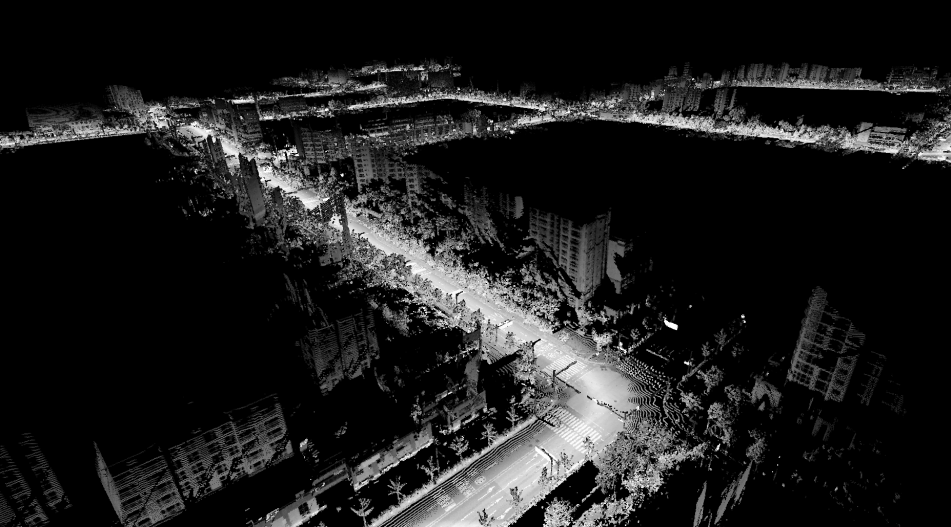}
    \label{fig:data_sample1}
  } \\    
  \caption{Point cloud sample data from 3D \ac{LiDAR} data and 2D \ac{LiDAR} data. Two different \ac{LiDAR} provide different aspect of the urban environment.}
  \label{fig:sample_pc}
\end{figure*}

%% file: conclusion.tex
\section{Conclusion and Future Work} 
\label{sec:conclusion}

This paper provided challenging data set targeting extremely complex urban environments where \ac{GPS} signals are not reliable. \bl{The data set provided a baseline generated using \ac{SLAM} algorithms with meter level accuracy.} The data sets also offer two-levels of sensor pairs for attitude and position. Commercial-grade sensors are less expensive and less accurate, while sensors such as \ac{FOG} and \ac{VRS} \ac{GPS} are more accurate and can be used for verification. The data sets captured various urban environments with different levels of complexity such as \figref{fig:sample_pc}, from metropolitan areas to residential areas. 

Our future data sets will be continually updated and the baseline accuracy will be improved. The future plan is to enrich the data set by adding a front stereo camera rig for visual odometry and a 3D \ac{LiDAR} to detect surrounding obstacles.